\documentclass[runningheads]{llncs}

\usepackage{ltl}
\usepackage{graphicx}
\usepackage{wrapfig}
\usepackage{subcaption}
\usepackage{listings}
\usepackage{array}
\usepackage{graphicx}
\usepackage{hyperref}
\usepackage{tikz}
\usepackage{listings}
\usepackage{xcolor}
\usepackage{xspace}
\usepackage{cite}
\usepackage{amsmath,amssymb,amsfonts}
\usepackage{algorithmic}
\usepackage{textcomp}
\usepackage[T1]{fontenc}
\usepackage{verbatim}
\usepackage{float}

\usepackage{lineno}

\def\BibTeX{{\rm B\kern-.05em{\sc i\kern-.025em b}\kern-.08em
    T\kern-.1667em\lower.7ex\hbox{E}\kern-.125emX}}

\usetikzlibrary{matrix}
\usetikzlibrary{positioning}
\usetikzlibrary{shapes.geometric, arrows}
\usetikzlibrary{shadows.blur}
\usetikzlibrary{tikzmark}
\usetikzlibrary{calc}
\tikzset{
  -|-/.style={->,draw,
    to path={
      (\tikztostart) -| ($(\tikztostart)!#1!(\tikztotarget)$) |- (\tikztotarget)
      \tikztonodes
    }
  },
  -|-/.default=0.5,
  |-|/.style={
    to path={
      (\tikztostart) |- ($(\tikztostart)!#1!(\tikztotarget)$) -| (\tikztotarget)
      \tikztonodes
    }
  },
  |-|/.default=0.5,
}

\newcommand{\upd}[2]{\ensuremath{[ \hspace{0.5pt} #1 \leftarrowtail \, #2 \hspace{0.5pt} ]}}
\newcommand{\name}[1]{{\text{\texttt{#1}}}}

\newcommand{\term}{\ensuremath{\tau}}
\newcommand{\fterm}{\ensuremath{\term_{F}}\xspace}
\newcommand{\pterm}{\ensuremath{\term_{P}}\xspace}

\newcommand{\terms}{\ensuremath{\mathcal{T}}\xspace}
\newcommand{\pterms}{\ensuremath{\terms_{\!P\hspace{-0.5pt}}}\xspace}
\newcommand{\fterms}{\ensuremath{\terms_{\!F\hspace{-0.5pt}}}\xspace}

\newcommand{\sep}{\ensuremath{\quad | \quad}}

\newcommand{\sats}{\ensuremath{\;\vDash_{\!\langle \hspace{-1pt} \cdot \hspace{-1pt} \rangle}}}

\newcommand{\branch}[2]{\ensuremath{#1 \! \wr \hspace{-0.6pt} #2}}
\newcommand{\assign}[1]{\ensuremath{\langle #1 \rangle}}
\newcommand{\functions}{\ensuremath{\mathcal{F}}\xspace}
\newcommand{\fnames}{\ensuremath{\mathbb{F}}\xspace}

\lstdefinelanguage{JavaScript}{
  morekeywords=[1]{break, continue, delete, else, for, function, if, in,
    new, return, this, typeof, var, void, while, with},
  morekeywords=[2]{false, null, true, boolean, number, undefined,
    Array, Boolean, Date, Math, Number, String, Object},
  morekeywords=[3]{eval, parseInt, parseFloat, escape, unescape},
  sensitive,
  morecomment=[s]{/*}{*/},
  morecomment=[l]//,
  morecomment=[s]{/**}{*/}, 
  morestring=[b]',
  morestring=[b]"
}[keywords, comments, strings]

\definecolor{mediumgray}{rgb}{0.3, 0.4, 0.4}
\definecolor{mediumblue}{rgb}{0.0, 0.0, 0.8}
\definecolor{forestgreen}{rgb}{0.13, 0.55, 0.13}
\definecolor{darkviolet}{rgb}{0.58, 0.0, 0.83}
\definecolor{royalblue}{rgb}{0.25, 0.41, 0.88}
\definecolor{crimson}{rgb}{0.86, 0.8, 0.24}

\lstdefinestyle{JSES6Base}{
  backgroundcolor=\color{white},
  basicstyle=\ttfamily\scriptsize,
  breakatwhitespace=false,
  breaklines=false,
  captionpos=b,
  columns=fullflexible,
  commentstyle=\color{mediumgray}\upshape,
  emph={},
  emphstyle=\color{crimson},
  extendedchars=true,  
  fontadjust=true,
  frame=single,
  identifierstyle=\color{black},
  keepspaces=true,
  keywordstyle=\color{mediumblue},
  keywordstyle={[2]\color{darkviolet}},
  keywordstyle={[3]\color{royalblue}},
  numbers=left,
  numbersep=5pt,
  numberstyle=\tiny\color{black},
  rulecolor=\color{black},
  showlines=true,
  showspaces=false,
  showstringspaces=false,
  showtabs=false,
  stringstyle=\color{forestgreen},
  tabsize=2,
  title=\lstname,
  upquote=true  
}

\lstdefinestyle{JavaScript}{
  language=JavaScript,
  style=JSES6Base
}
\lstdefinestyle{ES6}{
  language=ES6,
  style=JSES6Base
}

\begin{document}

\title{Guiding LLM Temporal Logic Generation with Explicit Separation of Data and Control}

\author{William Murphy, Nikolaus Holzer\inst{1}, Nathan Koenig\inst{1}, Leyi Cui\inst{2}, Raven Rothkopf\inst{2}, Feitong Qiao\inst{1}, and Mark Santolucito\inst{2}}
\institute{Columbia University, New York NY, USA \and Barnard College, Columbia University, New York NY, USA}

\maketitle

\begin{abstract}
Temporal logics are powerful tools that are widely used for the synthesis and verification of reactive systems.
The recent progress on Large Language Models (LLMs) has the potential to make the process of writing such specifications more accessible.
However, writing specifications in temporal logics remains challenging for all but the most expert users.
A key question in using LLMs for temporal logic specification engineering is to understand what kind of guidance is most helpful to the LLM and the users to easily produce specifications.
Looking specifically at the problem of reactive program synthesis, we explore the impact of providing an LLM with guidance on the separation of control and data--making explicit for the LLM what functionality is relevant for the specification, and treating the remaining functionality as an implementation detail for a series of pre-defined functions and predicates.
We present a benchmark set and find that this separation of concerns improves specification generation.
Our benchmark provides a test set against which to verify future work in LLM generation of temporal logic specifications.

\end{abstract}

\section{Introduction}\label{sec:intro}
Linear temporal logic has proven useful in improving the safety and scalability of a variety of systems. It allows an author to specify formal, mathematical constraints on the behavior of a system, and programs to be generated or verified against it. This has allowed programs to be generated that would be too complex for maintainers directly to understand, and has allowed systems to provide mathematical guarantees in high-stakes environments. Adoption of linear temporal logic has been limited by the difficulty of composing specifications. In many cases, the specification is much easier to write than the generated program; however, we believe the specification can be made even easier. Large Language Models (LLMs) have revolutionized both code generation and natural language understanding (NLU). We propose a strategy to leverage these two strengths of LLMs in order to allow users to specify systems in natural language rather than the symbolic language of linear temporal logic. The natural language is then translated into an LTL specification, which the user can inspect or apply to further steps in a verification or synthesis pipeline.

The field of linear temporal logic (LTL), the formal specification of the behavior of programs, has seen tremendous progress in recent years. LTL aims to place logical, mathematically rigorous guarantees on the behavior of a system, for the purpose of verifying an existing system or generating a new one. 
A primary example of the successful use of LTL is the AMBA bus protocol~\cite{bloem2014parameterized}. 
Since then, the field has seen significant milestones in education~\cite{ma2023using}, FPGA game development~\cite{geier2019syntroids}, music~\cite{choi2021program}, and interactive animations~\cite{rothkopf2023towards}.

An important extension of LTL is Temporal Stream Logic (TSL)~\cite{finkbeiner2019temporal}, which adds to LTL data of arbitrary type (rather than just Boolean) and the notions of updates, predicates, and functions. 
TSL can be used to synthesize programs that satisfy the TSL specification for all possible implementations of the function and predicate terms.
TSL is flexible and powerful, and its precision can be valuable to high-stakes environments like medicine, energy, transportation, and defense. However, its rigorous treatment of problems can be abstract to users. TSL's notational logic is challenging for less experienced users, and even experienced users need to devote significant time and careful attention to each application. 
The logical specification is also  decoupled from the reasoning required for implementing the function and predicate terms. This separation puts the burden on the user to determine what details about a system are needed to verify its behavior; by contrast, when a user discusses an example with a colleague, the problem is usually stated in an informal, natural language way that includes logical constraints mixed with illustrative details and facts about the implementation. When it comes to write the TSL spec, the user needs to separate out these constraints and the TSL spec more closely resembles a mathematical paper, stripped of all commentary and with only the equations.
A further avenue for improvement in TSL is the ease of editing existing specifications. For most real applications, modifications need to be made after a first version, and processes that aids editing for a user, who may not even be the original author, will greatly increase the value of TSL. Moreover, many TSL specifications take place in the same domain, using the same hardware, function implementations, and conceptual constructs. These domain-specific TSLs would benefit from reusable frameworks that capture the semantics of the target domain. 

In 2023, LLMs were used by 92\% of U.S.-based developers in writing code. 
Previous work has attempted to leverage LLMs for the generation of temporal logic specifications~\cite{cosler2023nl2spec, liu2022lang2ltl}; however, we propose to extend this work in two directions. First, we generate a set of benchmarks using examples of significant size and complexity that would be interesting to the community, both as (1) validation for the practicality of generating real-world specifications, and as (2) illustrative examples and inspiration for further work. Second, we apply LLM specification generation to temporal stream logic. We believe that TSL can enhance generated specifications by providing powerful abstractions to serve complex domains.

We believe that our pipeline provides a natural and helpful structure to the TSL specification process. As it stands, TSL specifications, isolated from any context, are challenging to parse for anyone but the original author. The current solution has been to include a variety of unstructured commentary about the specification, before, after, or interspersed in comments. Our pipeline gives structure to this context, providing information in a predictable, natural language format that is easy for humans to read, and contains enough detail to define a rigorous specification.

The core contributions of this paper can be summarized as follows:
\begin{enumerate}
    \item We improve the usability of TSL specifications by proposing a pipeline that leverages LLMs for code generation.
    \item We propose a set of benchmarks to test the practicality of this pipeline.
    \item We observe LLMs are often able to generate correct specifications, and that making explicit the separation of data and control helps to increase the accuracy of LLM specification generation.
\end{enumerate}

In particular, our pipeline makes use of three components, each of which the user must specify: (1) a short, high-level natural language summary of the problem (2) a series of constraints on the system, stated in plain English, and (3) the names and signatures of the function and predicates terms to be used in the specification. We structured this approach around what we imagined to be the most natural and human-friendly way to describe a specification, as a paper or colleague might describe a system before writing a formal specification. 

\section{Motivating Example}\label{sec:motive}
To demonstrate the utility of our method, we will first provide an overview of the current development process for reactive systems synthesized from Temporal Stream Logic (TSL) specifications.
We then describe how LLMs can be used to expedite this process by generating implementations for a TSL specification's signals, predicates, function terms, and context-specific setup.

\begin{figure}
\begin{lstlisting}
always assume {
    [ball <- moveLeft ball] -> X (! rightmost ball);
    [ball <- moveRight ball] -> X (! leftmost ball);
    ! (leftmost ball && rightmost ball);
}
always guarantee {
    rightmost ball -> F [ball <- moveLeft ball ];
    leftmost ball -> F [ball <- moveRight ball ];
    (! leftmost ball && ! rightmost ball ) -> 
        F ([ball <- moveLeft ball ] || [ball <- moveRight ball ]);
    (leftmost ball && X (!leftmost ball)) -> ((! [ball <- moveLeft ball]) W rightmost ball);
    (rightmost ball && X (!rightmost ball)) -> ((! [ball <- moveRight ball]) W leftmost ball);
}
\end{lstlisting}
\caption{\texttt{ball.tsl} example TSL specification.}
\label{fig:ball_tsl}
\end{figure}

For the sake of this motivating example, imagine we are trying to build a reactive system that controls ball bouncing from left to right.
We can specify the behavior of this system using TSL, demonstrated in Fig~\ref{fig:ball_tsl}.
While understanding the exact semantics of this specification is not necessary for this example, the high-level functionality is that the ball should bounce back and forth between the left and right boundaries of the screen.
The \texttt{ball} signal captures the position of the ball at each time step.
When the ball reaches the \texttt{leftmost} boundary, it should \texttt{moveRight}, and when it reaches the \texttt{rightmost} boundary, it should \texttt{moveLeft}.

After formalizing the temporal behavior of the reactive system, we can then synthesize the controller code using our TSL synthesis tool.
A synthesized JavaScript code snippet representing the first state for the ball bouncing system is shown in Fig~\ref{fig:ball_js}.
The code is a straightforward sequence of \texttt{if} statements with a global \texttt{currentState} variable, as well as any cell values that are instantiated as global variables.
\begin{figure}
\begin{lstlisting}[style=JavaScript]
if (currentState === 0) {
    if (!leftmost(ball) && rightmost(ball)) {
        ball = moveLeft(ball)
        currentState = 1
    }
    else if (!leftmost(ball) && !rightmost(ball)) {
        ball = moveLeft(ball)
        currentState = 1
    }
    else if (leftmost(ball) && !rightmost(ball)) {
        ball = moveRight(ball)
        currentState = 2
    }
}
\end{lstlisting}
\caption{\texttt{ball.js} synthesized code snippet from \texttt{ball.tsl}.}
\label{fig:ball_js}
\end{figure}

After synthesizing the controller code, we must implement predicates like \texttt{leftmost(), \texttt{rightmost()}} and function terms like \texttt{moveLeft(), moveRight()}, as well as the \texttt{ball} signal and the wrapper code for the animation to fully construct the reactive system.
The current standard practice is to manually implement the reactive system as one would normally--as a web-app, for example--and then refactor the code into pure functions and predicates that support the synthesized controller code.
This process is laborious and error-prone, as it requires developers to understand the synthesized code and the TSL specification to ensure that the system behaves as expected.
Moreover, this process must be repeated for every new feature or modification to the TSL specification, making it difficult to scale the development of reactive systems synthesized from TSL specifications.



\section{Background}\label{sec:bg}
What follows is a description of Temporal Stream Logic (TSL) and the TSL synthesis pipeline. 
We give only basic overview of TSL as it relates to the synthesis pipeline discussed in Sec.~\ref{sec:system} of the paper. 
For a full exposition of TSL's formal background, we refer the reader to prior work~\cite{finkbeiner2019temporal}.

\subsection{Reactive synthesis and Temporal Stream Logic (TSL)}
The goal of our system is to generate Temporal Stream Logic specifications based on natural language. These specifications describe systems that are reactive in the sense that they run in infinite loops, consuming input from the environment and producing output. 
In the classical synthesis setting, using specification languages such as Linear Temporal Logic~\cite{pnueli1977temporal}, time is discrete and inputs and outputs are given as vectors of Boolean signals. 
\emph{Temporal Stream Logic (TSL)}~\cite{finkbeiner2019temporal} is a specification language that provides an extra layer of abstraction that can be applied to input and output values of arbitrary type (as opposed to only Boolean as in LTL). 
Specifically, TSL introduces predicate terms, $ \pterm \in \pterms $, which are used to make observations on the environment, and function terms, $ \fterm \in \fterms $, which are used to construct output values.
These predicate terms allow users to separate the data and control of a system, encapsulating functionality in functions and predicates that is irrelevant to the specification, so the specification need only deal with the parts of the system that need to be understood in order to provide the desired guarantees. Additionally, TSL introduces the notion of cell values - values which are outputted by the system and piped back as input in the following timestep.
The grammar of a TSL formula, $\varphi$, is an extension to the grammar of LTL:
\begin{equation*}
  \begin{array}{rl}
  \pterm \ \;:= & \name{p}~\;\fterm^{0}~\;\fterm^{1}~\;\ldots~\;\fterm^{n-1} \\[0.3em]
  \fterm \ \;:= & \name{s}_{\name{i}} \sep \name{f}~\;\fterm^{0}~\;\fterm^{1}~\;\cdots~\;\fterm^{n-1} \\[0.3em]
  \varphi \ \; := & \; \pterm \ \, | \ \, \upd{\name{s}_{\name{o}}}{\fterm} \ \, | \ \, \neg \varphi \ \, | \ \, \varphi \wedge \varphi \ \, | \ \, \LTLnext \varphi \ \, | \ \, \varphi \LTLuntil \varphi
  \end{array}
\end{equation*}
where $ \name{s}_{\name{i}} \in \mathbb{I} \cup \mathbb{C}$ is an input stream or cell value, and $ \name{s}_{\name{o}} \in \mathbb{O} \cup \mathbb{C} $ is an output stream or cell value. 
Together, all the available predicate names $\name{p}$ and all the available function names $\name{f}$ form the set of function symbols $\fnames$.
A TSL formula describes a system that consumes input $\mathcal{I} = \mathbb{I} \cup \mathbb{C}$ and produces output $\mathcal{O} = \mathbb{O} \cup \mathbb{C}$.

\subsection{TSL Synthesis and Universal Quantification}
The realizability problem of TSL is stated as: given a TSL formula~$ \varphi $, is there
a strategy~$ \sigma \in \mathcal{I}^{+} \to \mathcal{O} $ mapping a finite input stream (since the beginning of time) to an output (at each particular timestep),
such that for any infinite input stream $ \iota \in \mathcal{I}^{\omega} $,
and every possible interpretation of the function symbols (some concrete implementation) $ \assign{\cdot} : \fnames \to \functions $, 
the execution of that strategy over the input $ \branch{\sigma}{\iota} $ satisfies $ \varphi $, i.e.,
\begin{equation*}
  \exists \sigma \in \mathcal{I}^{+} \to \mathcal{O}. \ \, 
  \forall \iota \in \mathcal{I}^{\hspace{0.2pt}\omega}. \ \, 
  \forall \assign{\cdot} : \fnames \to \functions. \ \, 
  \branch{\sigma}{\iota}, \iota \sats \varphi
\end{equation*}
If such a strategy~$ \sigma $ exists, we say that $ \sigma $ realizes $ \varphi $.
The key insight here is that in TSL we specify a temporal relation of predicate evaluations to function applications--abstracting away from what these predicates and functions actually do to any underlying data.
In TSL synthesis, this model $\sigma$ can be turned into a block of program code that describes a Mealy machine~\cite{mealy1955method}, where the transitions represent function and predicate terms.

Critically, TSL separates control from data, meaning that the synthesized controller code that realizes some TSL specification must be valid for any implementation $ \assign{\cdot} $ of the function and predicate terms; data manipulations are \emph{universally quantifiable}.
The universal quantification of TSL's function and predicate is particularly well-suited for the context of LLM specification generation.
This separation enables developers to first focus on when their system should execute certain behaviors, and leave the question of how those behaviors should be implemented for a later step in the development process.
Traditionally, the control is synthesized from TSL and the data is implemented manually by the end-user, but we propose to automate both processes using LLMs.

\section{System Overview}\label{sec:system}

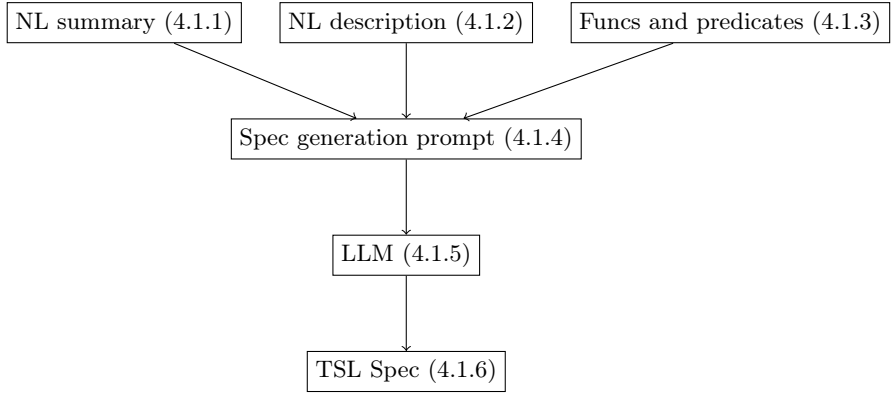
\begin{figure}[ht]
\centering
\begin{tikzpicture}[node distance=1cm and 0.5cm, auto]
  \node (nl_summary) [draw] {NL summary (4.1.1)};
  \node (nl_description) [draw, right=of nl_summary] {NL description (4.1.2)};
  \node (func_pred_terms) [draw, right=of nl_description] {Funcs and predicates (4.1.3)};
  
  \node (spec_gen_prompt) [draw, below=of nl_description] {Spec generation prompt (4.1.4)};
  \node (llm) [draw, below=of spec_gen_prompt] {LLM (4.1.5)};
  \node (tsl_spec) [draw, below=of llm] {TSL Spec (4.1.6)};
  
  \draw[->] (nl_description) -- (spec_gen_prompt);
  \draw[->] (nl_summary) -- (spec_gen_prompt);
  \draw[->] (func_pred_terms) -- (spec_gen_prompt);
  
  \draw[->] (spec_gen_prompt) -- (llm);
  \draw[->] (llm) -- (tsl_spec);
\end{tikzpicture}
\caption{The specification generation pipeline}
\label{fig:system}
\end{figure}

Our system is a prompt generation pipeline shown in Fig.~\ref{fig:system} with an ablation step to test removing several inputs. The core of the specification generation is to feed a specification (4.1.4) into a large language model (4.1.5), which then produces a TSL specification (4.1.6). 

The purpose of our system is to test the impact of including different pieces of information in the specification generation prompt and observing a change in the output specification. In particular, we include three important pieces of information:

\begin{enumerate}
    \item A high-level, natural language summary of the problem ("NL summary")
    \item A more detailed, natural language description of the most important assumptions and guarantees needed in the spec ("NL description")
    \item A separation of data and control in the form of function and predicate terms. The specification should use these to encapsulate logic not relevant to the assumptions and guarantees. ("Functions and predicates")
\end{enumerate}

\tikzstyle{box} = [rectangle, minimum width=3cm, minimum height=1cm, text centered, draw=black, fill=white]
\tikzstyle{arrow} = [thick,->,>=stealth]

As an example, consider the ``bouncing ball'' motivating benchmark. Below are the three inputs included in the prompt. The natural language summary is quite short, but paints a clear picture for the LLM of the abstractions involved. It is important to establish a clear metaphor in this ``high level summary'', because the LLM needs to be able to tie keywords together in subsequent natural language. For example, when we use words like ``wall'', ``bounce'', ``leftmost'', and ``move away'', the LLM will need to translate these natural language descriptions into a geometric understanding of the situation.

\begin{figure}[ht]
\centering
\begin{minipage}{0.9\textwidth}
\begin{verbatim}
A ball is bouncing between two walls. The position of the ball can
be represented as a number between 0 and 10, where 0 represents
that the ball is against the leftmost wall and 10 represents
that it is against the rightmost wall.
\end{verbatim}
\end{minipage}
\caption{Natural language summary}
\label{fig:nlsummary}
\end{figure}

The natural language description is the full description of the problem, after the summary. Below is an excerpt of the first few lines of a natural language description for the ball scenario. Here specific assumptions and guarantees are spelled out, but in plain English using the scenario described in the summary. When vocabulary is introduced in the summary (e.g. "Ball"), the LLM is capable of connecting vocabulary in the natural language description (e.g. "move away from") in a consistent abstraction of the scenario. It is not necessary to spell out what English terms mean, as long as the English terms all describe a consistent problem that a human reader would find plausible.

\begin{figure}[ht]
\centering
\begin{minipage}{0.9\textwidth}
\begin{verbatim}
Assumptions:

1. If the ball is moved to the left, then in the next moment, it is
guaranteed that the ball is not at the rightmost wall.

...

\end{verbatim}
\end{minipage}
\caption{Natural language description (excerpt)}
\label{fig:nldesc}
\end{figure}

The third part of the specification is a core contribution of our work. We separate the data and control of the problem by specifying an abstraction that the specification will use to encapsulate logic not relevant to the guarantees and assumptions. Here, we define functions ``moveLeft/moveRight'' and predicates ``leftmost/rightmost'' that move or determine the position of the ball. By defining these function interfaces, we cause the LLM to treat their logic as an irrelevant implementation detail, and to focus on the logic necessary for the guarantees.

\begin{figure}[ht]
\centering
\begin{minipage}{0.9\textwidth}
\begin{verbatim}
Cells:
  "ball" is a cell that represents the state of the signal that
  determines how the ball should move
Functions:
  moveLeft(ball) => returns a signal to move the ball to the left
  moveRight(ball) => returns a signal to move the ball to the right
Predicates:
  leftmost(ball) => is the ball against the leftmost wall?
  rightmost(ball) => is the ball against the rightmost wall?
\end{verbatim}
\end{minipage}
\caption{Functions and predicates description}
\label{fig:funcandpred}
\end{figure}

\section{Evaluation}\label{sec:eval}
In this section, we present an evaluation on the efficacy of separating data and control for generating TSL specifications with LLMs.
We expand upon our motivating example in Sec~\ref{sec:motive} by evaluating our approach on a series of benchmarks we devised for this purpose.
This shows when the separation of data and control improves the specification generation process.

\subsection{Benchmarks}

We evaluated the pipeline on a series of benchmarks designed to test a number of situations in which data was separated from control. 
In particular, we ran four classes of benchmarks, and made three variations on one of the classes (Cube) in order to test changes to the behavior of the generated program based on different assumptions and guarantees in the same environment\footnote{Our full set of benchmarks is publicly available at 
\url{https://github.com/Barnard-PL-Labs/TSL_LLM_Benchmark}}.
These benchmarks aim to capture a range of simple reactive systems that describe visual systems.

We examined the quality of the generated specifications in three ways:

\begin{enumerate}
    \item How often did the LLM produce a valid specification?
    \item How often did the specification accurately model the problem?
    \item Given that the specification was correct, how often did it accurately model the problem?
\end{enumerate}

We considered (3) important because it is possible to generate several specifications and automatically detect and exclude malformed specifications. Below is a table of the results for each of the benchmarks, evaluated according to each of the three measures of quality.

\begin{figure}[h]
\centering 
\includegraphics[width=4.8in]{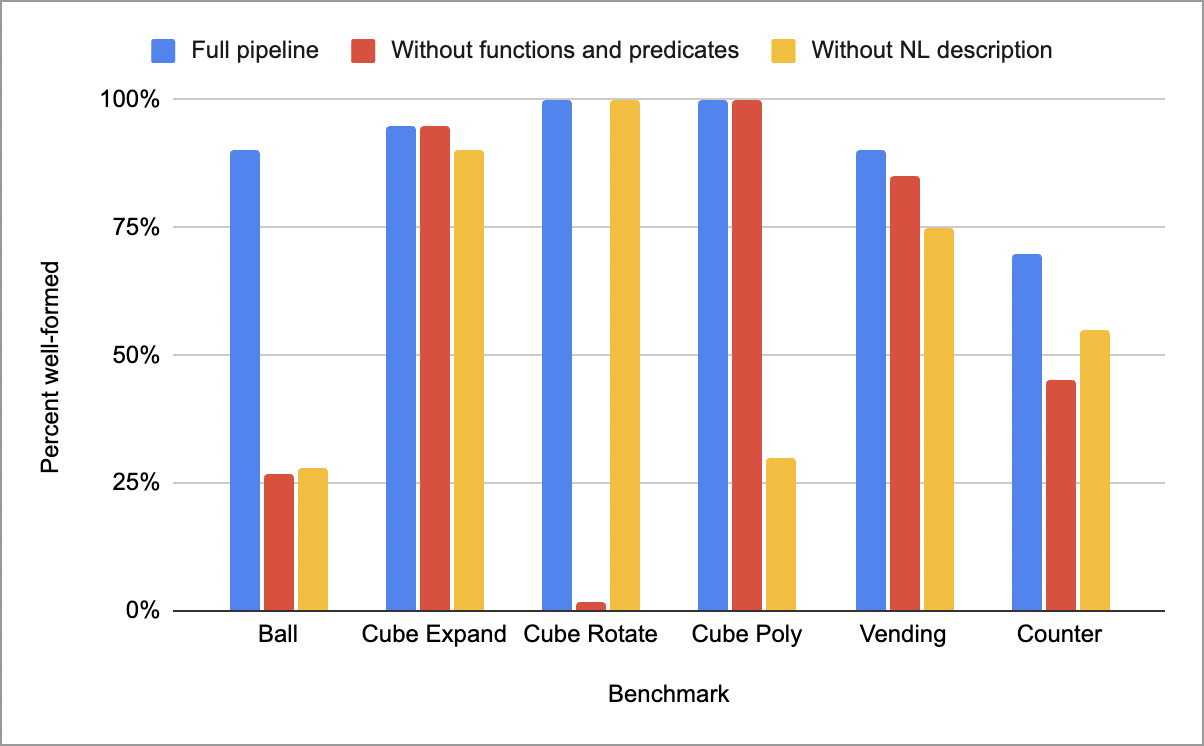} 
\caption{Pipeline accuracy, measured by (1) incidence of valid specification}
\label{fig:compiled}
\end{figure}

\begin{figure}[h!]
\centering 
\includegraphics[width=\textwidth]{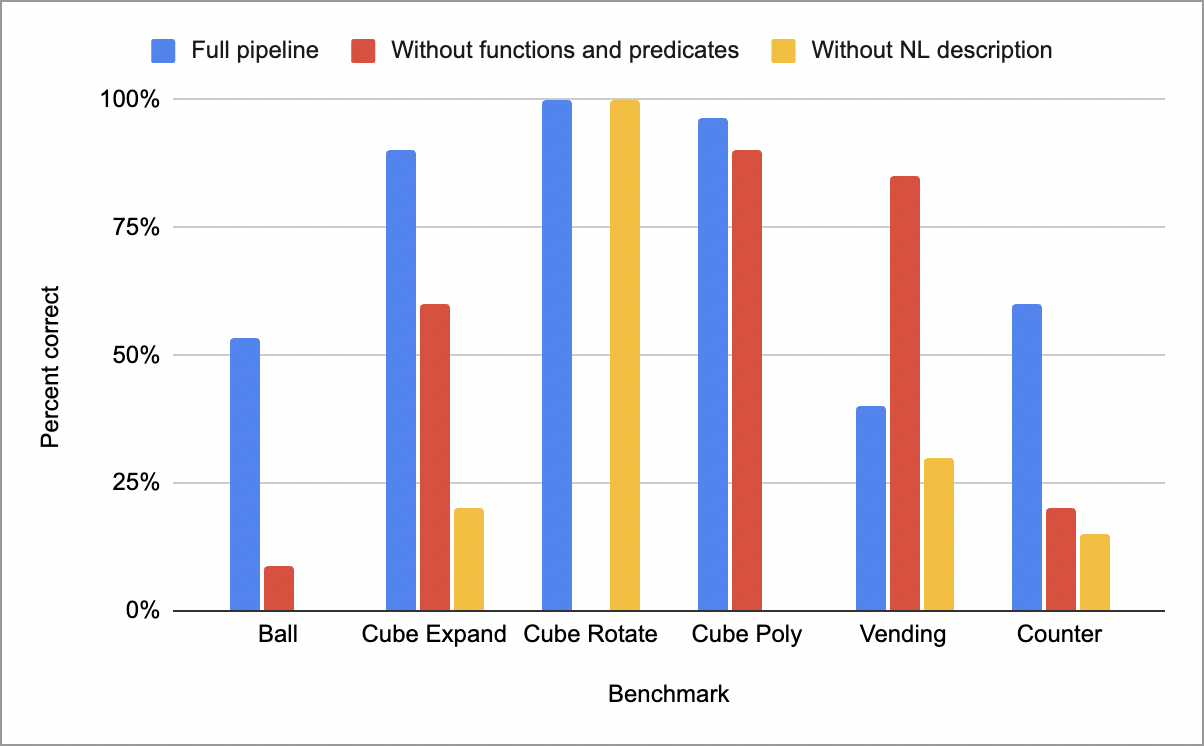} 
\caption{Pipeline accuracy, measured by (2) incidence of correct specification} 
\label{fig:correct} 
\end{figure}

\begin{figure}[h!]
\centering 
\includegraphics[width=\textwidth]{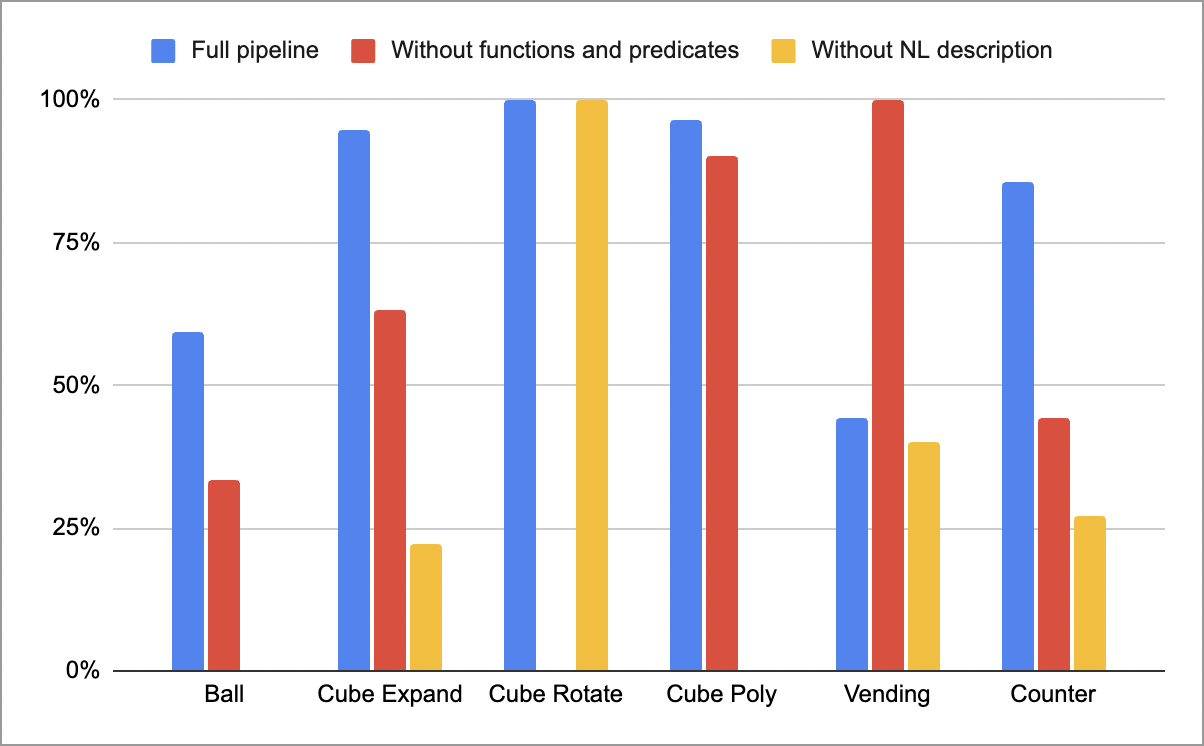} 
\caption{Pipeline accuracy, measured by (3) incidence of correctness given valid specification} 
\label{fig:correct_over_compiled} 
\end{figure}

Our results that the versions of the pipeline that separates data and control through function terms tended to outperform ones without functions and predicates, but not always. The benefit of this separation tended to disappear when the LLM seemed not to understand the definition of the function and predicate terms. In particular, the LLM would sometimes treat the function and predicate terms as though they had a different definition from what we intended, and this caused problems. For example, in the benchmark "Vending", the LLM misinterpreted the definition of "isLessThanPoint75", causing it to incorrectly subtract off negative numbers.

While further investigation is necessary, these results give initial evidence that using LLM for temporal specification generation is 

\section{Related Work}\label{sec:related}

LLM Enabled Code Generation
Large Language Models (LLMs) have quickly moved code generation to tackle problems that were previously far out of scope for traditional program synthesis techniques~\cite{swebench}. However there is still a need for software generation that has can provide formal guarantees of correctness. 
A key aspect of such guarantees is having formal language descriptions (i.e. specifications) of software.

In pursuit of LLM assisted specifciation generation, recent works, such as NL2spec~\cite{cosler2023nl2spec} and Lang2LTL~\cite{liu2022lang2ltl}, have explored the transformation of natural language descriptions into LTL specifications. These approaches aim to bridge the gap between informal user requirements and formal temporal logic specifications, enabling a more accessible and intuitive method for defining system behaviors and properties.

Further down the pipeline, we might imagine our LLM generated specifications being used for the automatic generation of systems. Prior work has explored using reactive synthsis (and TSL specifications specifically) as a method for LLM content generation~\cite{rothkopf2024enforcing}. Specifically, Rothkopf et al. explore how reactive synthesis can be leveraged to enforce temporal constraints on content generated by LLMs. 

\section{Conclusion}\label{sec:eval}

In conclusion, our goal was to test the value of separating control logic and data for improving LLM generation of temporal logic specifications. We used Temporal Stream Logic to separate control and data via function and predicate terms, and we devised a pipeline that incorporated a predefined set of abstractions before generating the specification. We created a set of benchmarks and performed an ablation study, removing functions and predicates as well as all natural language description except for a high-level summary. Our results indicate that a clear abstraction improves specification generation, especially when the abstraction is chosen in such a way that it is easy for the LLM to understand. This benchmark set provides a useful framework for future work, including problems hard enough to be useful even after significant improvement in LLM capabilities.


\bibliographystyle{splncs04}
\bibliography{refs}
\end{document}